\documentclass[final]{cvpr}

\usepackage{times}
\usepackage{epsfig}
\usepackage{graphicx}
\usepackage{amsmath}
\usepackage{amssymb}
\usepackage{booktabs}
\usepackage{bm}
\usepackage{verbatim}
\usepackage{enumitem}
\usepackage{colortbl}
\usepackage{xcolor}
\usepackage{multirow}

\usepackage[pagebackref=true,breaklinks=true,colorlinks,bookmarks=false]{hyperref}

\def\cvprPaperID{7516} % *** Enter the CVPR Paper ID here
\def\confYear{CVPR 2021}

\begin{document}

%%%%%%%%% TITLE
\title{Nothing But Geometric Constraints: A Model-Free Method for Articulated Object Pose Estimation}

\begin{comment}

\author{Qihao Liu\\
Johns Hopkins University\\
{\tt\small qliu45@jhu.edu}
% For a paper whose authors are all at the same institution,
% omit the following lines up until the closing ``}''.
% Additional authors and addresses can be added with ``\and'',
% just like the second author.
% To save space, use either the email address or home page, not both
\and
Second Author\\
Institution2\\
{\tt\small secondauthor@i2.org}
}
\end{comment}

\author{Qihao Liu,
\quad Weichao Qiu,\quad Weiyao Wang,\quad Gregory D. Hager,\quad Alan L. Yuille\\
Johns Hopkins University\\
{\tt\small \{qliu45,\ wwang121\}@jhu.edu,\ hager@cs.jhu.edu,\ \{qiuwch,\ alan.l.yuille\}@gmail.com }
}
\maketitle

\newcommand{\blue}[1]{\textcolor{blue}{#1}}

%%%%%%%%% ABSTRACT
\begin{abstract}
   We propose an unsupervised vision-based system to estimate the joint configurations of the robot arm from a sequence of RGB or RGB-D images without knowing the model a priori, and then adapt it to the task of category-independent articulated object pose estimation. We combine a classical geometric formulation with deep learning and extend the use of epipolar constraint to multi-rigid-body systems to solve this task. Given a video sequence, the optical flow is estimated to get the pixel-wise dense correspondences. After that, the 6D pose is computed by a modified P$n$P algorithm. The key idea is to leverage the geometric constraints and the constraint between multiple frames. Furthermore, we build a synthetic dataset with different kinds of robots and multi-joint articulated objects for the research of vision-based robot control and robotic vision. We demonstrate the effectiveness of our method on three benchmark datasets and show that our method achieves higher accuracy than the state-of-the-art supervised methods in estimating joint angles of robot arms and articulated objects.
\end{abstract}

%%%%%%%%% BODY TEXT
\section{Introduction}
\label{intro}
\begin{figure}
    \centering
    \includegraphics[width=0.47\textwidth]{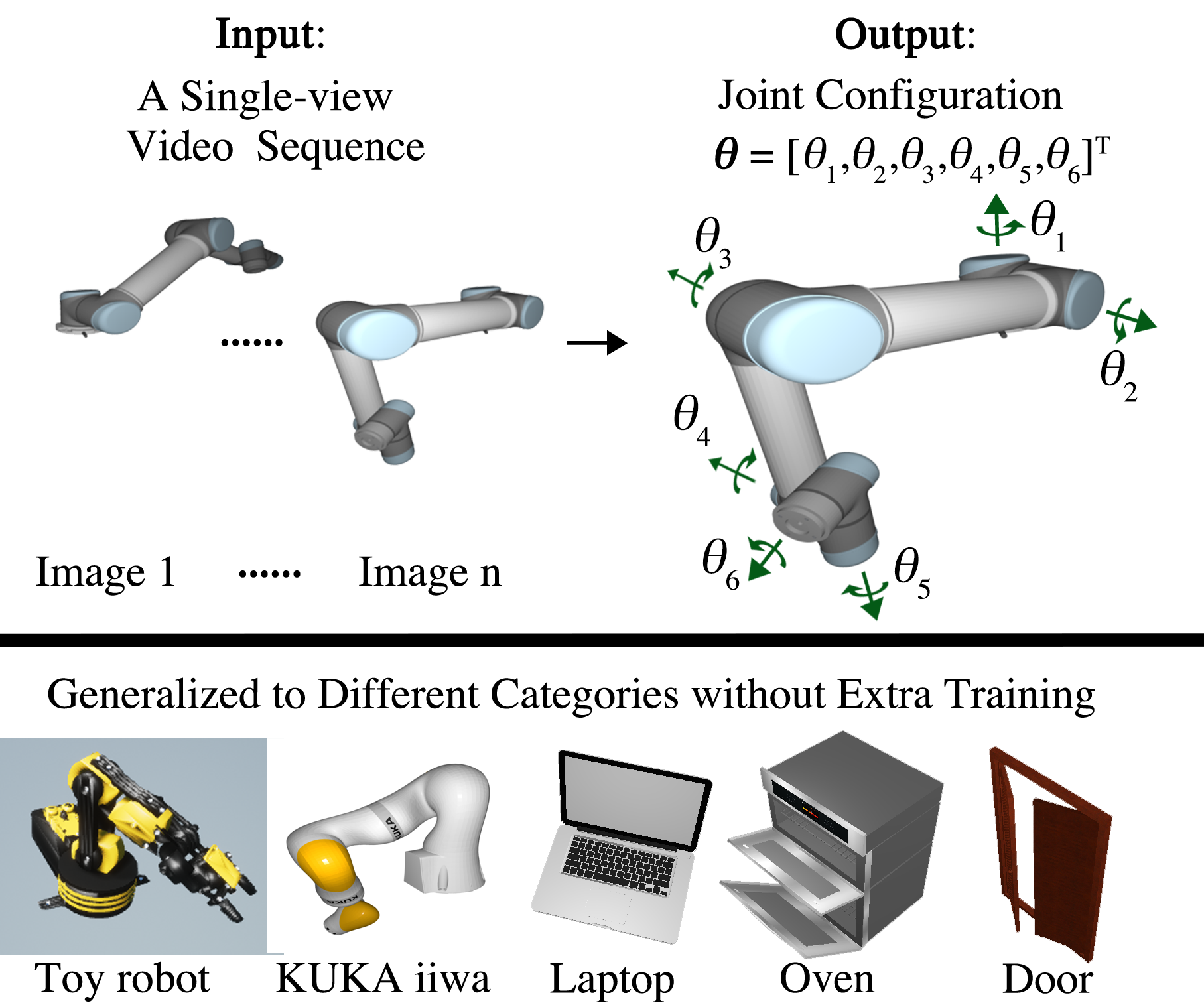}
    \caption{\textbf{GC-Pose} estimates the joint configurations of articulated objects from a video sequence. It is an unsupervised model-free method, thus it can be generalized to novel objects. The robot arm can be seen as a kind of articulated object with more degrees of freedom (DoF) and a more complicated motion pattern.}
    \label{fig:teaser}
    \setlength{\belowcaptionskip}{-3cm} 
\end{figure}

Articulated object pose estimation is becoming popular due to its high application value. It can enable operating real-world articulated objects. It can also be used to control a low-cost robot arm~\cite{zuo2019craves}, provide a safe-net for sensor failure~\cite{magrini2020human}, and allow collaboration between multiple arms~\cite{ha2020learning}. Solving this problem can significantly increase the ability of a robot to interact with the real world.

Most recent articulated object pose estimation methods are instance-level or category-level, which means the model needs to be trained with data of the same object~\cite{loper2015smpl} or from the same object category  \cite{li2020category}. 
The requirement of training data limits applicable scenarios. 
Instance-level methods are fragile when coming across a novel object. 
Category-level methods suffer from large intra-class variation. 
However, when humans see a novel articulated object, we can discover moveable parts of it by observing its video, and estimate the joint configurations of these parts. 
This makes us wonder whether our method can estimate the articulated object pose without a CAD model of the object. Without knowing the object shape beforehand, can we achieve similar performance to the supervised models?

% Fit a model into the image, video? The SE-net, the talk paper, the SLAM paper.
% Can we compare with previous model free approach? Can we say previous model-free is not popular?

In this paper, we propose an unsupervised model-free method, namely \textbf{GC-Pose}, to estimate the joint configurations of any kind of robot arms or articulated objects from a single-view video sequence. We are particularly interested in the joint configurations $\mathbf{q}$ of the articulated object (or joint angles $\bm{\theta}$ for a robot arm) since its pose is well-determined given the joint configurations. Given a video sequence, our method first detects the articulated parts by estimating the dense correspondence between frames using optical flow and clustering the motion vectors. Then we compute the 6D transformation of each moving part from the 2D correspondence. Unlike the supervised models, which directly give the absolute joint configuration, our method first estimates the relative transformation between frames. Then the absolute joint configuration is computed by accumulated the relative motion over frames.

Model-free methods suffer from many challenges. 
To overcome these challenges and achieve high accuracy, we introduce some novel components in our method. Challenges of articulated object pose estimation include self-occlusion, low-textured area, reflective surface, ambiguous pose, tiny parts, etc. 
First, the low-textured and reflective surface of an object makes the dense correspondence noisy. 
To address this challenge, we extend the epipolar constraint from the multi-view scene to a single-view multi-rigid-body scene and use it with the EM algorithm \cite{dempster1977maximum} to refine the part-level optical flow.
%The epipolar constraint is taken from multi-view vision, but in our case, the camera is fixed, while the moving object is the object part.
This constraint holds since each moving part is a rigid body and the optical flow for this part should be satisfied with one 6D transformation.
We combine the epipolar constraint with the EM algorithm to estimate the rigid-body optical flow at the part level and get a more accurate optical flow estimate despite the existence of low-textured area.
Second, to handle the ambiguous pose and self-occlusion, we use bundle adjustment to optimize the results over multiple frames to find the best pose estimate from the video.

Experiments show our method can largely surpass existing model-free models and achieve similar performance compared with supervisely trained models. We also use synthetic data to perform ablation study to understand the properties of our model.
We evaluate the performance of our system by conducting three sets of experiments on three benchmark datasets.
%  including a novel one with different kinds of robots and multi-joint articulated objects. 
% state-of-the-art (SOTA) 
We compare our system with ScrewNet~\cite{jain2020screwnet}, a model-free method, and with CRAVES~\cite{zuo2019craves}, a supervised robot pose estimation algorithm.
% The experimental results show that our system outperforms these supervised pose estimation methods. 
Then we perform ablation studies on our synthetic dataset. 
Our synthetic data helps us to study the impact of our modules by controlling image appearance and generating intermediate ground truth.
Our experiments demonstrate that leveraging the geometric constraints of the rigid body and the constraint between multiple frames helps handle the low-textured area and complicated motion pattern, which leads to improved performance in unsupervised, model-free articulated pose estimation.

In summary, our contributions are:
\begin{itemize}[topsep=0pt,itemsep=0pt,parsep=0pt]
\item We propose a model-free pose estimation method (GC-Pose), which can estimate the joint configurations of robot arms or articulated objects from a video sequence without CAD model or annotated dataset.
\item We develop the \textit{extended epipolar constraint} and show that it improves pose estimation on low-textured areas of a rigid body.
% \item We develop a benchmark dataset of different kinds of robots and multi-joint articulated objects with various labels for vision-based robot control, robotic vision, and robotic manipulation tasks.
\item We perform extensive experiments to show the effectiveness of our method on both simulated and real data. We also develop a synthetic dataset focusing on controllability to analyze the properties of our model.
\end{itemize}

\section{Related work}
%This section summarizes related work on robot arm pose estimation and articulated object pose estimation, considering robot arm as articulated objects. We also mention some related work on synthetic data for robots.
%\cite{byravan2017se3}
%\cite{schmidt2014dart}
%\cite{zuo2019craves}
%\cite{vijayanarasimhan2017sfm}

\subsection{Vision-based Robot Arm Pose Estimation}

Robot pose estimation has been studied extensively for a long time. Many previous methods need to know the exact robot model\cite{bohg2014robot,widmaier2016robot,ortenzi2016vision} or need fiducial markers\cite{ortenzi2018vision}. For example, Bohg et al.\cite{bohg2014robot} use a random decision forest to segment the link of the robot, then estimate its joints. Widmaier et al.\cite{widmaier2016robot} propose a pixel-wise regression-based method to directly regressing to the robot joint configurations.

Recently, the keypoint-based methods are used in robot pose estimation\cite{miseikis2018multi,zhou20193d,lambrecht2019towards,gulde2018ropose,gulde2019ropose,ma2020edge}. A keypoint detector is trained in these methods to estimate the keypoints or joint positions on a specific robot arm, which needs a large amount of real data with labels. Zuo et al.\cite{zuo2019craves} also propose a keypoint-based network trained using synthetic data, with domain adaptation to bridge the reality gap. A key difference is that our method doesn't need the CAD model or trained on labeled data, we use dense correspondence from optical flow instead of keypoints to estimate the joint angles.

Different from the previous methods estimating the joint configuration, Lee et al.\cite{lee2020camera} propose a keypoint-based method to estimate the camera-to-robot pose. Byravan et al.\cite{byravan2017se3} explore the use of SE(3) transformation to learn the rigid body motion. It takes 3D point clouds as input and predicts the 3D optical flow of the scene. A key difference is that our method computes optical flow from the 2D image sequence and estimates the 3D rigid body motion represented by joint angles. 

\begin{comment}
\cite{bohg2014robot} -- pixel-wise part classification, need cad model x
\cite{widmaier2016robot} -- pixel-wise regression-based, need cad model x
\cite{ortenzi2016vision} -- robot part tracking with CAD model x
\cite{ortenzi2018vision} -- Keypoint based using marker, KUKA x
\cite{ma2020edge} -- Keypoint based, Rokae robot arm x
\cite{miseikis2018multi} -- CNN end-to-end,  UR5 x
\cite{zhou20193d}\cite{lambrecht2019towards} -- Keypoint based, UR5 x
\cite{gulde2018ropose}\cite{gulde2019ropose} -- synthetic data+real data, keypoint based, UR5 x
\cite{lee2020camera} -- use keypoint and P$n$P to estimate the camera extrinsics, assuming the joint configuration is known. x
\cite{zuo2019craves} x
\end{comment}

\subsection{Articulated Object Pose Estimation}
Articulated object pose estimation is the prerequisite for robot manipulating daily objects with functional parts. As a key for robots to move out of factories into homes and our daily environment, articulated object pose estimation is attracting more and more attention. One line of work uses a robot manipulator to interact with the articulated objects, then use the 3D points to infer the pose\cite{katz2013interactive,katz2008manipulating,martin2016integrated}. These methods can perform pose estimation for unknown objects, but it takes a lot of time for interaction and can only be applied to simple articulated objects. Kumar et al. \cite{kumar2016spatiotemporal} estimate joint articulated motion in a standard simultaneous localization and mapping (SLAM) pipeline, but it can only be used to track a single landmark with 1-DoF joint. Schmidt et al. \cite{schmidt2014dart,schmidt2015depth} focus on tracking articulated bodies or objects with probabilistic inference, but formally defined geometric structures are required. Daniele et al. \cite{daniele2020multiview} incorporate both vision and natural language information for articulation model estimation. However, it needs natural language descriptions of motion as an additional observation mode.

Other approaches assume that instance-level information \cite{michel2015pose,ye2014real,desingh2019factored,pavlasek2020parts} or category-level information \cite{abbatematteo2019learning,yi2018deep,li2020category} is available. For example, Michel et al.\cite{michel2015pose} use a random forest to vote for pose parameters for each point in a depth image. Pavlasek et al. \cite{pavlasek2020parts} formulate the problem of articulated object pose estimation as a Markov Random Field. However, these approaches are limited by the need to have exact CAD models or object mesh models. Abbatematteo et al. \cite{abbatematteo2019learning} train a mixed density model on depth images and infer the kinematic model using probabilities prediction of a mixture of Gaussians. Li et al. \cite{li2020category} extend the NOCS \cite{wang2019normalized} to accommodate articulated objects at both part and object level. However, these methods assume that the category-level information is available.  Recently, Jain et al. \cite{jain2020screwnet} use screw theory to estimate the pose of an articulated object from a sequence of depth images without requiring prior knowledge of the articulation model category. However, this approach can only handle articulated models with at most one DoF, and labeled data is required for training.

\begin{figure*}
    \centering
    \includegraphics[width=\textwidth]{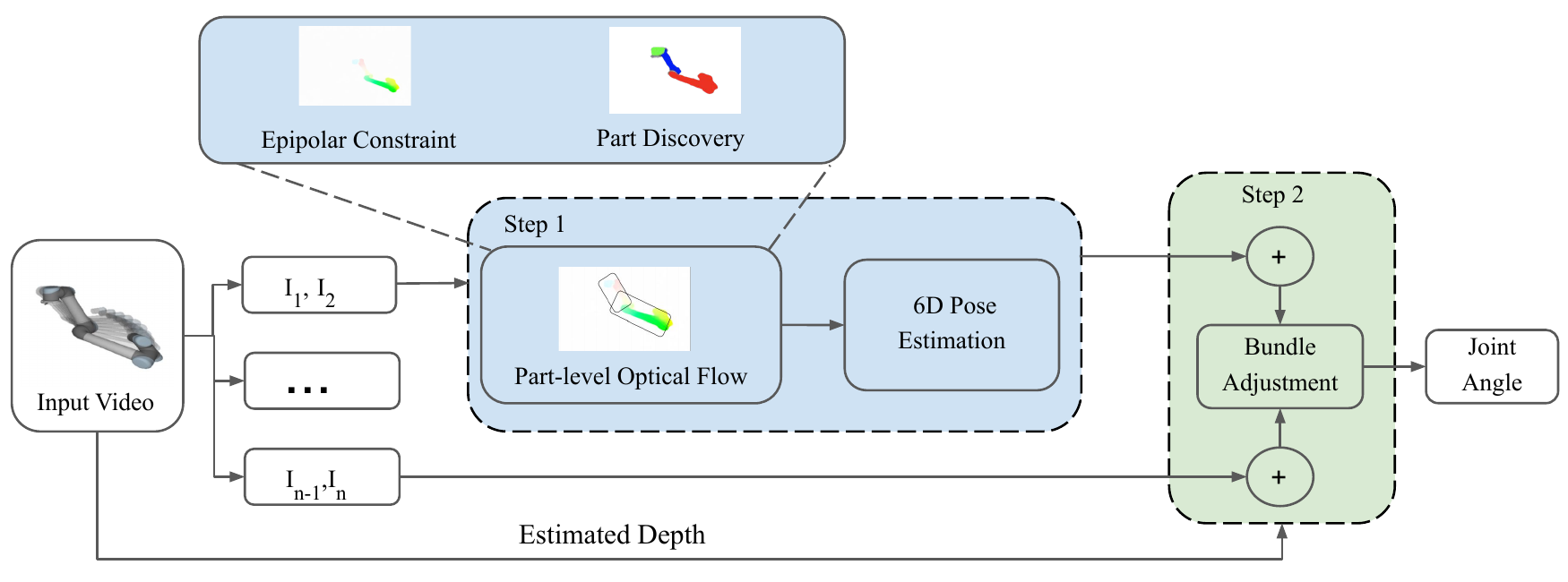}  
    \caption{\textbf{Schematic view:} The problem is solved in two steps. First, we estimate the change of the joint configurations between adjacent two frames. Then, we accumulate all the estimated configurations together and use bundle adjustment to optimize the results.}
    \label{fig:pipeline}
\end{figure*}

\section{Method}
In this section, we introduce our method and the synthetic dataset. In Sec.\ref{System Overview}, we first give an overview of our method. Then we describe three main components of our method. Next, we introduce our synthetic dataset in Sec.\ref{dataset}. Lastly, we provide some implementation details in Sec.\ref{implementation}.

%\subsection{Problem statement}

\subsection{System Overview}
\label{System Overview}
We first formulate the problem here. The ultimate goal is to estimate the joint configurations of a robot arm or any other articulated object from a single-view video without knowing the CAD model a priori.  Mathematically, given a sequence of $n$ images $\mathcal{I}_{1},...,\mathcal{I}_{n}$, we want to estimate the change of the joint configurations $\mathbf{q}_{1:n}$ in the given sequence of images. Considering that most of the robot arms or robot hands (e.g., UR5, KUKA, and Shadow hand) are composed of multiple revolute joints, we are particularly interested in joint angles $\bm{\theta}_{1:n}=[\theta^1_{1:n},\theta^2_{1:n},...,\theta^d_{1:n}]^\text{T}$, where $d$ denotes the DoF of the robot arm. We assume the DoF of the robot and the camera parameters are known.

The general idea to solve this problem is that, given a video sequence $\mathcal{I}_{1:n}$, the first step is to compute the change of joint angles $\bm{\theta}_{k:k+1}=[\theta^1_{k:k+1},\theta^2_{k:k+1},...,\theta^d_{k:k+1}]^\text{T}$ between the adjacent two frames $\mathcal{I}_{k}$ and $\mathcal{I}_{k+1}$, and the second step is to accumulate them to get the change of the joint angles in the entire video sequence, i.e., $\bm{\theta}_{1:n}=\sum_{k=1}^{n-1}\bm{\theta}_{k:k+1}$. Fig.\ref{fig:pipeline} shows the pipeline of our pose estimation system. 

In the first step, we compute the change of the joint angles $\bm{\theta}_{k:k+1}$. We first estimate the part-level optical flow from adjacent two frames $\mathcal{I}_{k}$ and $\mathcal{I}_{k+1} $(Sec. \ref{epiploss}). Then, for each rigid part $\alpha$ ,$\alpha\in \{1,...,d\}$, we use the estimated optical flow to compute the 6D pose transformation $\mathbf{T}_k^\alpha=[\mathbf{R}_k^\alpha|\mathbf{t}_k^\alpha]$ and compute the rotation angle $\theta^\alpha_{k:k+1}$ from the rotation matrix $\mathbf{R}_k^\alpha$ (Sec. \ref{3D2D}). In the second step, to obtain accurate results when accumulating the joint angles $\bm{\theta}_{k:k+1}$, two challenges needed to be addressed. One is that the optical flow estimate is noisy and the joint angles computed from the optical flow are not accurate, thus accumulating these results causes larger errors. The other challenge is the ambiguity of this problem caused by the complicated motion pattern of the object. To address these challenges, we utilize the constraint between multiple frames and use bundle adjustment to optimize the joint angles before accumulating them together to get the final estimate (Sec. \ref{BA}). 

\subsection{Part-level Optical Flow from Epipolar Constraint}
\label{epiploss}
In this subsection, we first describe the epipolar constraint in multi-rigid-body systems and then introduce the combination of the EM algorithm to optical flow method to get a more accurate part-level optical flow estimate.

\textbf{Epipolar constraint in multi-rigid-body systems:} The epipolar constraint is widely used in classical geometric methods\cite{hartley2003multiple,chen2019self}. Traditionally, it is an algebraic relationship that couples two or more 3D scene projections using a fundamental or essential matrix. Specifically, given two pixels $\mathbf{p}$ and $\mathbf{p}'$ in two different views, the traditional epipolar constraint defines their geometric relationship via the camera intrinsics and geometric of the camera displacement. This geometric relationship can be extended to a multi-rigid-body scene even though the camera remains still. 

Assume that $\mathbf{x}$ and $\mathbf{x}'$ are the 3D corresponding points of $\mathbf{p}$ and $\mathbf{p}'$ in the camera coordinate frame $\mathcal{C}$. For simplicity, and in a slight abuse of notation, we use $\mathbf{p}$ and $\mathbf{x}$ as both the homogeneous and non-homogeneous coordinates in this paper. Then, for rigid body $\alpha$, ${\mathbf{x}^\alpha}'=\mathbf{R}^\alpha \mathbf{x}^\alpha+\mathbf{t}^\alpha$, we have 
%where $\alpha$ denotes the part label, $\mathbf{x}^\alpha$ denotes the 3D point belongs to the rigid part $\alpha$, $\mathbf{R}^\alpha$ and $\mathbf{t}^\alpha$ denote the rotation matrix and translation vector of the rigid part $\alpha$, respectively. Considering that ${\mathbf{x}^\alpha}'\cdot(\mathbf{t}^\alpha\times \mathbf{x}^\alpha)=0$, we have 
% This is standard; you don't need to derive -- just take the result and move on
$({\mathbf{x}^\alpha}')^\text{T}[\mathbf{t}^\alpha]_\times\mathbf{R}^\alpha\mathbf{x}^\alpha=0$, where $\mathbf{x}^\alpha$ denotes the 3D points belong to the rigid part $\alpha$ and $[\mathbf{t}^\alpha]_\times$ is the skew-symmetric matrix of the translation vector $\mathbf{t}^\alpha$. Thus for each pixel $\mathbf{p}$ of rigid body $\alpha$, we get the following algebraic relationship:
\begin{align}
    ({\mathbf{p}^\alpha}')^\text{T}\mathbf{K}^{-\text{T}}[\mathbf{t}^\alpha]_\times\mathbf{R}^\alpha\mathbf{K}^{-1}\mathbf{p}^\alpha=0
\end{align}
 where $\mathbf{K}$ is the camera intrinsics. It defines the geometric relationship between two pixels via the camera intrinsics and geometric of the rigid body displacement. Therefore, we can incorporate this epipolar constraint as a penalty over the dense correspondences via the following loss function:
\begin{align}
    \mathcal{L}_e=\sum_\alpha||[\textbf{p}^\alpha+\textbf{F}_k(\textbf{p}^\alpha)]^\text{T}\textbf{K}^{-\text{T}}[\textbf{t}_k^\alpha]_\times\textbf{R}_k^\alpha\textbf{K}^{-1}\textbf{p}^\alpha||
\end{align}
where $\mathbf{p}^\alpha$ denotes the pixels belong to rigid body $\alpha$, $\mathbf{R}_k^\alpha$ and $\textbf{t}_k^\alpha$ are the 3D rotation matrix and translation vector of rigid body $\alpha$ from image $\mathcal{I}_k$ to $\mathcal{I}_{k+1}$, respectively.
%Notice that in this function, $\alpha$ is a hidden variable. 
This loss function enforces an epipolar constraint on the optical flow of each rigid part in a single view scene. 

\textbf{Part-level optical flow:} To address the challenge caused by the low-textured area and obtain a more accurate optical flow estimate, we combine the EM algorithm with optical flow method to estimate the rigid-body optical flow at the part level.
We use the pre-trained official FlowNet2 model \cite{ilg2017flownet} here. In our method, computing the epipolar loss needs to know the part label $\alpha$ for each pixel, while the part label is assigned based on the epipolar loss. Therefore, the epipolar constraint is used alternately in fine-tuning the pre-trained FlowNet2 model and in motion-based image segmentation. Specifically, given adjacent two images $\mathcal{I}_k,\ \mathcal{I}_{k+1}$ and the optical flow $\mathbf{F}_k(\cdot)$ estimated by the current FlowNet2 model, we first use RANSAC \cite{fischler1981random} to initialize the part label $\alpha$. Then we use the EM algorithm to optimize the label assignment for each pixel by minimizing the epipolar loss $\mathcal{L}_e$ computed based on the current optical flow estimate $\mathbf{F}_k(\cdot)$. After the EM algorithm converges, we fine-tune the current FlowNet2 model by minimizing a weighted sum of the epipolar loss that computed based on the current label of each pixel and the photometric difference loss which defined as followed:
\begin{align}
    \mathcal{L}_p=||\mathcal{I}_k(\mathbf{p})-\mathcal{I}_{k+1}(\mathbf{p}+\mathbf{F}_k(\mathbf{p}))||
\end{align}
Then we use the updated FlowNet2 model to estimate the optical flow and learn the new part label $\alpha$ for each pixel. This iteration continues until the part label for each pixel converges. The use of epipolar constraint enforces rigid body constraints on the optical flow. It implies that the optical flow of each rigid part should be able to be explained as a rigid body motion in 3D space. It can be used to improve the optical flow model on scenes with multiple rigid bodies and handle the low-textured area of a rigid body.

\subsection{6D Pose Transformation from Optical Flow}
\label{3D2D}
Given the estimated part-level optical flow $\mathbf{F}_k(\mathbf{p}^\alpha)$, the pixel-wise dense correspondences $(\mathbf{p}^\alpha,\mathbf{p}^\alpha+\mathbf{F}_k(\mathbf{p}^\alpha))$ is computed. Then we can use the dense correspondence to compute the transformation matrix. To avoid overly complicated notation, we ignore the subscript $k$ and superscript $\alpha$ here and denote the corresponding pixels by $(\mathbf{p},\mathbf{p}')$. 

We first compute the transformation matrix w.r.t to the camera coordinate frame $\mathcal{C}$. As mentioned in Sec. \ref{epiploss}, for the 3D corresponding points $\mathbf{x}$ and $\mathbf{x}'$, we have $(\mathbf{x}')^\text{T}[\mathbf{t}]_\times\mathbf{R}\mathbf{x}=0$, where $\mathbf{R}$ and $\mathbf{t}$ denotes the 3D rotation matrix and translation vector, respectively. They can be solved by running the eight-point algorithm \cite{szeliski2010computer} and then singular value decomposition. This step is similar to structure from motion (SfM), but the difference is that the objects are moving but the camera remains still in our problem.
%You should add cites for all the algorithms (e.g. Szelinski's book). Also, FWIW, the 8 point algorithm is not the best algorithm to use here ... there are much better approaches. But, if you bundle adjust it probably doesn't matter.

Then, notice that the transformation matrix $\mathbf{R}$ is defined w.r.t. the camera coordinate frame $\mathcal{C}$ (i.e., $\mathbf{R}$ should be written as $^\mathcal{C}{\mathbf{R}}$ ), we need to transform it into the model coordinate frame $\mathcal{M}$ and compute $^\mathcal{M}{\mathbf{R}}$ using the following equation: $^\mathcal{M}{\mathbf{R}}=(\ ^\mathcal{C}{\mathbf{R}}_{\mathcal{W}}\  ^\mathcal{W}{\mathbf{R}}_{\mathcal{M}})^\text{T}\ ^\mathcal{C}{\mathbf{R}}(\ ^\mathcal{C}{\mathbf{R}}_{\mathcal{W}}\  ^\mathcal{W}{\mathbf{R}}_{\mathcal{M}})$
, where $\mathcal{W}$ denotes the world coordinate frame, $\ ^\mathcal{C}{\mathbf{R}}_{\mathcal{W}}$ is the camera extrinsics, and $^\mathcal{W}{\mathbf{R}}_{\mathcal{M}}$ can be computed using kinematic chain and other estimated joint configurations. We can thus compute the pose transformation $\ ^\mathcal{M}{\mathbf{R}}$ from 2D correspondences and then use it to compute the joint angle $\theta^\alpha_{k:k+1}$ with the fact that each revolute joint has only one DoF.

\subsection{Bundle Adjustment}
\label{BA}
 
After the rotation angle $\bm{\theta}_{k:k+1}$ between adjacent two frames is computed, we utilize the constraint between multiple frames and use bundle adjustment to optimize the joint angles $\bm{\theta}_{k:k+1}$ before accumulating the results together to get the final estimate. Specifically, we first estimate the depth $\mathbf{D}_k(\mathbf{p})$ from the adjacent two frames $\mathcal{I}_{k}$ and $\mathcal{I}_{k+1}$. Then we compute the point cloud model $\mathbf{X}_k$ given by the 2D pixels $\mathbf{p}$ and depth $\mathbf{D}_k(\mathbf{p})$. Next, we select the 3D inliers for rigid body $\alpha$ that satisfy the following inequality:
\begin{align}
    \mathcal{L}_w^\alpha = ||[\mathbf{p}^\alpha+\mathbf{F}_k(\mathbf{p}^\alpha)]-\mathbf{K}\cdot \mathbf{T}^\alpha_k\cdot \mathbf{X}^\alpha_k|| < t
    \label{eq1}
\end{align}
where $\mathbf{T}_k^\alpha$ is the transformation matrix computed from the rotation angle $\bm{\theta}_{k:k+1}$ and $t$ is a threshold. After that, we optimize the transformation matrix $\mathbf{T}_k^\alpha$ by minimizing the loss function $\mathcal{L}^\alpha_w$ defined as the left-hand side of the inequality \ref{eq1}. Therefore, for each adjacent two frames $\mathcal{I}_{k}$ and $\mathcal{I}_{k+1}$, $k=1,...,n-1$, we can get a noisy point cloud model $\mathbf{X}^\alpha_k$ and the optimized transformation matrix $\mathbf{T}^\alpha_k$ which can be used to compute the joint angle for each rigid body. We accumulate both the joint angles $\bm{\theta}_{k:k+1}$ and the point cloud models $\mathbf{X}_k$. Finally, after going through the entire video sequence, we use the accumulated point cloud model $\mathbf{X}_{1:n-1}$ to optimize the joint angle $\bm{\theta}_{1:n}$ and get the final estimate. 

\subsection{Dataset}
\label{dataset}

Existing robot simulators focus on providing a virtual environment for accomplishing tasks.
They are powerful but difficult for vision researchers to learn and use.
The physics simulation and planning is not essential for 3D vision research. So we build our own simple and flexible rendering pipeline for generating synthetic data of articulated objects.

Our tool is built on top of the pyrender\footnote{\url{https://github.com/mmatl/pyrender}} project. We extend it by adding ground truth generation, texture manipulation, and domain randomization. 
The ground truth, including depth and optical flow, is used to analyze and understand our model in the ablation study.

% Our synthetic data focuses on controllability, ground truth, intermediate data for analyzing models. depth, optical flow.

Our tool can load the Unified Robot Description Format (URDF) and GL Transmission Format (glTF). The URDF format is widely used to define the geometry of a robot. It is used by robotic arm manufacturers and by researchers to create articulated object datasets~\cite{xiang2020sapien,li2020category}. The glTF format is a standard for exchanging 3D models. 
URDF format is suitable for modeling articulated objects with rigid parts, but it's difficult to model a human body, in which the mesh of body part deforms during the animation. The glTF format has no such limitations.
Its nice integration with industry tools makes it a standard for model creators to share their models on the internet.
We can find many science fiction robots (like R2D2, Transformers, Gundam, etc.) shared on the internet with this format. This greatly enriches the set of articulated objects which we can test on.

% \blue{More technical details are going to be added here.}

% pyrender gltf format and URDF format, one is commonly for robot arm definition. The other is for other animatable object, such as human or a quad-leg robot.

% Change texture for abalation study and creating background.

% provide a python API to change texture, generate optical flow through dense correspondence.

\subsection{Implementation Details}
\label{implementation}
\textbf{Network:} Our approach involves estimating optical flow and estimating depth (if the input is a sequence of RGB images). To estimate optical flow, we use the pre-trained official FlowNet2 model \cite{ilg2017flownet} and fine-tune the small displacement network and fusion network of FlowNet2 model with learning rate $\gamma = 5\times 10^{-7}$.  We set the weight coefficients of both epipolar loss and photometric loss equal to 1. The model is fine-tuned for 50 epochs on a single NVIDIA TITAN X GPU with a batch size of 1. Other hyper-parameters are the same as in the original paper \cite{ilg2017flownet}. To estimate depth, we use the pre-trained official Monodepth2 model \cite{godard2019digging}.

\textbf{Bundle Adjustment:} When using the accumulated point cloud model $\mathbf{X}_{1:n-1}$ to optimize the joint angle $\bm{\theta}_{1:n}$, the cost function we minimized is different from the cost function we used when optimizing the joint angle $\bm{\theta}_{k:k+1}$ since we don't have dense correspondence here. Therefore, we optimize the joint angle $\bm{\theta}_{1:n}$ by minimizing the following cost functions:
\begin{align}
    \mathcal{L}_b^\alpha = ||\mathcal{I}_n(\mathbf{p}^\alpha) - \mathcal{I}_n(\mathbf{K}\cdot\mathbf{T}_{\theta^\alpha_{1:n}}\cdot\mathbf{X}^\alpha_{1:n-1})||
\end{align}
where $\mathbf{T}_{\theta^\alpha_{1:n}}$ denotes the transformation matrix of joint angle $\theta^\alpha_{1:n}$. We optimize the joint angle via gradient descent algorithm with a learning rate $\gamma_{ba}=10^{-3}$.

\section{Experiments}
\begin{figure*}
    \centering
    \includegraphics[width=\textwidth]{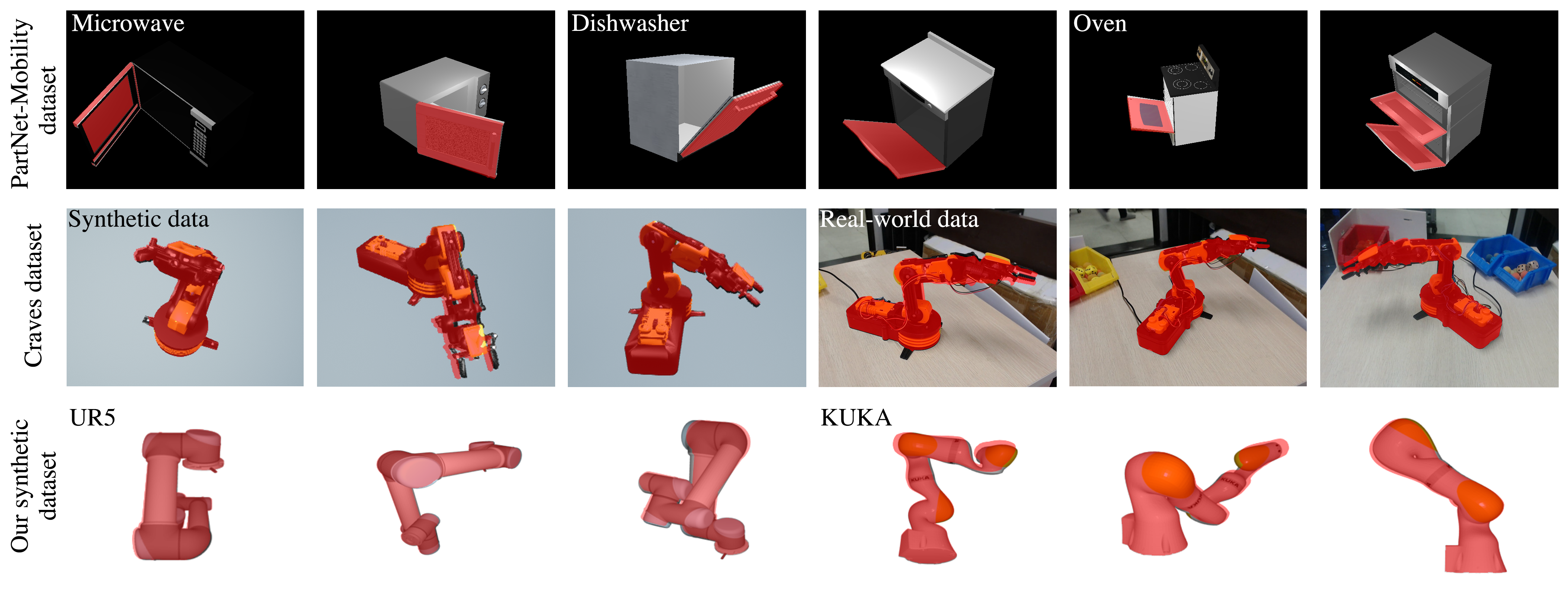}  
    \caption{\textbf{Qualitative results}: 
    The results are presented as overlapped images, in which the original object and the red one represent the ground-truth pose and the estimated pose, respectively. The top row shows test results on the object instances from different categories of the PartNet-Mobility dataset \cite{xiang2020sapien} (for microwave, dishwasher, and oven category, respectively). The middle row shows test results on the toy robot arm from the dataset provided by Zuo et al. \cite{zuo2019craves}. The first three show the results on synthetic data while the last three show the results on the real-world data. The bottom row shows test results on the UR5 and KUKA robot arms from our synthetic dataset. To better visualize the results, we remove the background for the synthetic data and only the target object remains. Notice that the background images are added during the evaluation.}
    \label{fig:Qualitative results}
\end{figure*}

We evaluate the performance of the proposed method in estimating the joint angles of robot arms and articulated objects by conducting three sets of experiments on three benchmark datasets. In Sec. \ref{exp_set}, we first introduce the datasets and the baseline algorithms. Then we compare our method with baselines in Sec. \ref{com_with_baseline}. Finally, we conduct ablation studies in Sec. \ref{Abl} to demonstrate the effectiveness of the constraints we exploited in the proposed method. The qualitative results are shown in Fig. \ref{fig:Qualitative results}
\subsection{Experimental Setup}
\label{exp_set}
\textbf{Datasets:}
To evaluate our method on the task of articulated object pose estimation, we use the recently proposed PartNet-Mobility dataset \cite{xiang2020sapien} that contains 46 kinds of articulated objects. We follow the baseline algorithms \cite{jain2020screwnet} and evaluate our method on the microwave, dishwasher, and oven category, each of which contains 1000 images.
To evaluate our method on the task of robot arm pose estimation, we generate 1600 synthetic video sequences with randomized camera parameters, arm motions, and background. Each video sequence includes 15 images. Specifically, we generate 1000 video sequences for the UR5 robot arm, 200 video sequences for the UR5 robot with textured surface, 200 for the KUKA iiwa robot arm, and 200 for the ABB IRB 2400 robot arm. We also generate 200 video sequences ($\sim 3000$ images) for the OWI-535 robot arm following the same rendering pipeline provided by Zuo et al. \cite{zuo2019craves}. 

\textbf{Baselines:}
Our method can be used to solve the task of category-independent articulated object pose estimation and robot arm pose estimation. Therefore, we first compare the proposed method with the SOTA supervised category-independent articulated object pose estimation method named ScrewNet \cite{jain2020screwnet}. We also compare our method with the articulation model estimation method proposed by Abbatematteo et al.\cite{abbatematteo2019learning} and the Iterative Closest Point algorithm (ICP) \cite{besl1992method}. When testing the ICP baseline, we use the estimated optical flow with ground-truth depth to estimate 3D dense correspondences and report error only about the axis of rotation for comparison. 
Moreover, We compare the proposed method with the SOTA supervised robot arm pose estimation method named Craves \cite{zuo2019craves}.

\subsection{Comparison with baselines}
\label{com_with_baseline}
\begin{table*}
    \small 
    \centering
    \begin{tabular}{ccccc}
    \toprule
    &Microwave&Dishwasher&Oven&Oven (Across object class)\\
    \midrule 
    ScrewNet \cite{jain2020screwnet}&$2.549\pm0.939^{\circ}$&$4.037\pm1.613^{\circ}$&$\mathbf{0.720\pm0.140^{\circ}}$&$8.650\pm0.207^{\circ}$\\
    ICP \cite{besl1992method}&$9.895\pm8.043^{\circ}$&$14.396\pm10.472^{\circ}$&$11.227\pm9.311^{\circ}$&$11.227\pm9.311^{\circ}$\\
    Abbatematte et al. \cite{abbatematteo2019learning}&$11.856\pm2.456^{\circ}$&$9.735\pm4.600^{\circ}$&$13.087\pm4.649^{\circ}$&$9.915\pm3.934^{\circ}$\\
    \rowcolor{gray!10}\textbf{GC-Pose}&$\mathbf{1.231\pm0.717^{\circ}}$&$\mathbf{1.605\pm1.057^{\circ}}$&$1.337\pm1.199^{\circ}$&$\mathbf{1.337\pm1.199^{\circ}}$\\
    \bottomrule
    \end{tabular}
    \caption{\textbf{Performance comparison of articulated object pose estimation on the PartNet-Mobility dataset} \cite{xiang2020sapien}. The results are presented as mean and standard diversion of the error, evaluated over 1000 images for each category.}
    \label{SAPIEN}
\end{table*}

\begin{table*}
    \small 
    \centering
    \resizebox{\textwidth}{13mm}{
    \begin{tabular}{cccccc}
    \toprule
    & \multicolumn{4}{c}{Synthetic data}&\multicolumn{1}{|c}{\multirow{2}{*}{Real-world data}}\\
    \cmidrule {2 -5}
    &Rotation&Base&Elbow&\multicolumn{1}{c|}{Wrist}&\\
    \midrule
    Craves\cite{zuo2019craves}&$2.8415\pm0.8263^{\circ}$&$3.0825\pm1.0282^{\circ} $&$2.8257\pm 0.9200^{\circ}$&$3.3806\pm1.2193^{\circ}$&$4.4128\pm 1.931^\circ$\\
    \rowcolor{gray!10}\textbf{GC-Pose} (RGB)&$2.9719\pm 1.2091^{\circ}$&$2.8177\pm 1.9490^{\circ}$&$3.0132\pm1.6279^{\circ}$&$4.1522\pm2.6377^{\circ}$&$\mathbf{3.1416\pm 2.850^\circ}$\\
    \rowcolor{gray!10}\textbf{GC-Pose} (RGB-D)&$\mathbf{1.7632\pm 1.0900^{\circ}}$&$\mathbf{1.3688\pm 1.0862^{\circ}}$&$\mathbf{1.7357\pm 1.3744^{\circ}}$&$\mathbf{2.9779\pm1.7250^{\circ}}$&-\\
    \bottomrule
    \end{tabular}}
    \caption{\textbf{Performance comparison of robot arm pose estimation on the synthetic dataset and real-word dataset (called Lab dataset) provided by Zuo et al} \cite{zuo2019craves}. The results are presented as mean and standard diversion of the error, evaluated over 3000 synthetic images and 381 real images. Since the real-world data only has 381 images, we average the error over all the joints instead of dividing the results by different joints and computing the mean error for each joint.}
    \label{CRAVE}
\end{table*}

%In this section, we consider the task of estimating articulated object poses and robot arm poses, and compare the performance of our method with the baseline methods in these tasks. 
\subsubsection{Category-independent Articulated Object Pose Estimation}

We compare our method with ScrewNet\cite{jain2020screwnet}, ICP \cite{besl1992method}, and the method proposed by Abbatematteo et al. \cite{abbatematteo2019learning} on the PartNet-Mobility dataset \cite{xiang2020sapien}. For this set of experiments, we use the model fine-tuned on our synthetic UR5 robot arm datasets and then evaluate it on 1000 images for each of the three categories from the PartNet-Mobility dataset. During the evaluation, we refine the results using epipolar constraint for 10 iterations and report the estimated joint angles and the error of the last iteration.

Table \ref{SAPIEN} shows the quantitative results of this set of experiments. Due to the same experiment condition, we directly use the reported results of ScrewNet. Different from our method which is an unsupervised method and doesn't need extra training on specific category, ScrewNet is a supervised method that can be trained on the images from the same object category or across object categories. Therefore, they report two different sets of experiments. The first three columns show the results of ScrewNet trained and tested on the same object class, and the last column shows the results of ScrewNet trained on the dishwasher and the microwave category but tested on the oven category.

As shown in table \ref{SAPIEN}, our method achieves significant gains over the baseline, as well as the other competing methods in the category of microwave and dishwasher. In the category of oven, ScrewNet performs very well when trained on the images from the same category. However, when trained on the images from other object classes, it performs much worse than our GC-Pose. If we compute the mean error of our GC-Pose across different categories, we can find that the mean error and the standard deviation of the error are $1.391^{\circ}$ and $0.157^{\circ}$, respectively. Therefore, our method is not only more accurate than the baselines but also more stable and generic than other algorithms since our method is model-free and its accuracy depends only on the motion pattern and the surface of the object.

\subsubsection{Robot Arm Pose Estimation}
We compare our method with Craves \cite{zuo2019craves} on the synthetic dataset generated following their provided rendering pipeline. Craves is the SOTA semi-supervised robot pose estimation method which trained on labeled synthetic as well as unlabeled real data, and obtained enough accuracy for vision-based robot control. Another reason we choose it as a baseline is that it focuses on low-cost robot arm control and enables closed-loop control for no-sensor robot arms. It is one of the most meaningful applications for robot pose estimation. Craves requires an exact CAD model but only needs RGB images as input. For a fair comparison, we conduct two groups of experiments to evaluate GC-Pose, one of which uses RGB images as input, and the other uses RGB-D images. No CAD model is used in our GC-Pose. We use the FlowNet2 model fine-tuned on our synthetic UR5 robot arm datasets. During the evaluation, we refine the results using epipolar constraint for 10 iterations and report the estimated joint angles and the error of the last iteration.

Table \ref{CRAVE} shows the quantitative results of this set of experiments. The first four columns show the results evaluated on the synthetic dataset. We report the angular error for each joint of the OWI-535 robot arm. For Craves, we directly use the pre-trained official Craves model and it produces an average error of $3.03^{\circ}$ on the synthetic data, which is similar to the reported average error in the original paper \cite{zuo2019craves}. As shown in Table \ref{CRAVE}, our GC-Pose produces an average error of $3.23^{\circ}$ when using RGB images as input and an average error of $1.96^{\circ}$ when using RGB-D images as input. It can be observed that our model-free method achieves comparable performance with the supervised baseline method when using RGB images but outperforms it when extra depth information is available. 

The last column of Table \ref{CRAVE} shows the results evaluated on the real-word data. Since the dataset only provide real-world images with no depth information, thus we only evaluate our GC-Pose with RGB images. We report the mean angular error for all the joints of the OWI-535 robot arm. Our method performs better than Craves on the real-world dataset even if no depth information is provided. Considering that an average error of $4.81^{\circ}$ is enough for vision-based control of the OWI-535 robot arm according to the original paper \cite{zuo2019craves}, we believe that our method can also be used in vision-based control of the same robot arm.
\subsection{Ablation Study}
\label{Abl}
\begin{table*}
    \small 
    \centering
    \resizebox{\textwidth}{18mm}{
    \begin{tabular}{cccccccccc}
    \toprule
    &&\multicolumn{2}{c|}{UR5}& \multicolumn{2}{c|}{UR5 (textured surface)}&\multicolumn{2}{c|}{KUKA iiwa}&\multicolumn{2}{c}{ABB IRB 2400}\\
    \cmidrule {3 -10}
    & &mean error&\multicolumn{1}{c|}{PCP@$5^{\circ}$}&mean error&\multicolumn{1}{c|}{PCP@$5^{\circ}$}&mean error&\multicolumn{1}{c|}{PCP@$5^{\circ}$}&mean error&PCP@$5^{\circ}$\\
    \midrule
    \multirow{3}{*}{\rotatebox{90}{RGB}}&FlowNet2 (FN) &$25.7216^{\circ}$&12.5\%&$8.4578^{\circ}$&$45.0\%$&$17.1803^\circ$&19.5\%&$29.1733^\circ$&$3.5\%$\\
    &FN+Epipolar Loss (EL) &$8.7170^{\circ}$&$49.2\%$&$5.6970^\circ$&$65.5\%$&$8.1617^\circ$&42.0\%&$9.0064^\circ$&$31.0\%$\\
    &FN+EL+Estimated Depth &$3.8168^{\circ}$&$85.9\%$&$3.1532^\circ$&$87.5\%$&$4.0264^\circ$&83.5\%&$5.9153^\circ$&$81.5\%$\\
    \cmidrule {1 -2}
    \multirow{2}{*}{\rotatebox{90}{RGBD}}&FN+EL+Gt Depth &$2.2132^{\circ}$&$91.6\%$&$1.4633^\circ$&$96.5\%$&$2.5158^\circ$&89.0\%&$3.1479^\circ$&$93.5\%$\\
    &FN+EL+Gt Depth w Noise &$2.9758^{\circ}$&$89.6\%$&$2.2413^\circ$&$94.0\%$&$3.2217^\circ$&$87.0\%$&$4.6583^\circ$&$85.0\%$\\
    \bottomrule
    \end{tabular}}
    \caption{\textbf{Ablation study:} We evaluate ablated versions of our GC-Pose on our synthetic dataset. We remove the videos that include robot collision, serious self-occlusion, or moving pattern cannot be estimated by the vision-based method. The results are presented as the mean angular error and the percentage of angles which are predicted within $5^{\circ}$ of the given ground-truth rotation angles (PCP@$5^\circ$).The first three columns use RGB images as input and the last two columns use RGB-D images as input.}
    \label{Ablation_Study}
\end{table*}

We consider four ablated versions of our method and evaluate them on our synthetic dataset. First, to test the effectiveness of epipolar constraint, we consider two ablated versions of our GC-Pose, one of which directly estimate the pose from the pre-trained official FlowNet2 model and the other estimate the posed from a fine-tuned FlowNet2 model using epipolar constraint. As the second ablation study, we focus on the depth information we used and compare the performance of our GC-Pose when using the estimated depth, the ground-truth depth, and the ground-truth depth with noise. Notice that we don't attach any end-effectors to the robot arm, and we only estimate the bottom three joints. The results are reported in Table \ref{Ablation_Study}. 

\subsubsection{Epipolar Constraint}
\textbf{Does epipolar constraint help handle the low-textured area?} From the first two rows of Table \ref{Ablation_Study}, we can find that the performance is significantly improved by using epipolar constraint. If we compare the results of the original UR5 and UR5 with a textured surface (i.e., the first two columns), we can find a gap between the performance in these two cases when we directly estimate the pose from the result of FlowNet2 (as shown in the first row). The estimated optical flow is more accurate on the textured surface, and thus it provides more precise point correspondences. However, when using epipolar constraint (i.e., the second row of the first two columns), we can see a similar performance on the UR5 with the textured and low-textured surface. 

\noindent\textbf{Why does it work?} This is understandable, as the epipolar constraint provides an implicit expression for the optical flow of each rigid body in the form of 6D transformation. This low DoF implicit expression enforces extra geometric constraints on the low-textured area with the idea that the optical flow of each rigid body should be able to be explained as a rigid body motion in 3D space. Therefore, we argue that the epipolar constraint can help handle the low-textured area of a rigid body and provide more accurate dense correspondences for further pose estimation.

\subsubsection{Depth Information}
\textbf{Does bundle adjustment improve results?} From the last four rows of Table \ref{Ablation_Study}, we can find that the use of bundle adjustment yields much better performance than the ablated version with only epipolar constraint (i.e., the second row of Table \ref{Ablation_Study}). The reason why bundle adjustment improves performance is that it can bound the estimated joint angles between adjacent two frames by introducing the 3D shape of the moving part. When accumulating the joint angles, if the estimated angles of any two adjacent frames produces a large error, it will cause a larger error in the final estimate. This can be observed from the second row of Table \ref{Ablation_Study} that $49.2\%$ of the error is smaller than $5^{\circ}$, but the mean error is $8.7170^{\circ}$, which is much bigger than $5^{\circ}$. The ablated version without bundle adjustment works well in some simple cases, but it fails in the other difficult cases in which self-occlusion and complicated motion patterns occur. This problem can be solved when we reconstruct a 3D shape and use the reconstructed shape to optimize the estimated results (as shown in the last three rows of Table \ref{Ablation_Study}).

\noindent\textbf{Is GC-Pose robust to the error of depth estimation?} Now, we look at the last three rows of Table \ref{Ablation_Study}. The third row reports the results when using the pre-trained official Monodepth2 model to estimate the depth. The fourth row shows the results with accurate ground-truth depth, which can be seen as the upper-bound or our method. The last row displays the results when using the ground-truth depth that contains a Gaussian distribution noise with a standard deviation of $5\ cm$.  It is observed that even if we use the estimated
depth, we can still obtain relatively accurate results. But apparently, the more accurate the depth information we use, the better estimate we can get. In brief, the use of depth information alongside standard RGB images can improve the performances, but it also restricts the real-world application. When the depth information is not available, the estimated depth can serve as an alternative.

\section{Conclusions}
We build an unsupervised model-free articulated object pose estimation system that only requires video sequences as input. It first leverages the geometric constraints to get a more accurate optical-flow estimate, then uses the 2D dense correspondence to estimate the 6D pose of each rigid body, and lastly, it uses bundle adjustment to optimize the joint parameters.  Compared to prior SOTA, our method is more accurate, easier to train, and generalizes better in solving the task of articulated object pose estimation. Our GC-Pose does not need the 3D CAD model or labeled training data and can be applied to different robot arms and different categories of articulated objects by several epochs of fine-tuning on the raw video sequence. Moreover, we build a synthetic dataset that includes various kinds of robots and multi-joint articulated objects. The dataset focuses on controllability and can be used to diagnose and analyze the properties of different methods for vision-based robot control, robotic manipulation, and pose estimation.

Our formulation opens many directions for future work. In particular, how to build a real-time end-to-end robot arm pose estimation system to provide joint configurations for further control, thus help robots move out of factories into daily environments. Moreover, We combine epipolar constraint with the EM algorithm and optical flow method, and our results are more accurate than the supervised method on standard benchmarks, suggesting a new direction of combining classical formulations with deep learning.

% Dataset
% Baseline  % SMPL, craves
% Abalation

\begin{comment}

Dataset: CRAVES dataset, our synthetic + real, what else?

Evaluation metric?

Closed loop control system

PCP@$5^{\circ}$: Percentage of Correct Pose, the percentage of angles which are predicted within $5^{\circ}$ of the given ground truth rotation angles.
\end{comment}
%With and without epipolar constraint

%With and without bundle adjustment

%With and without texture, control robot arm reluctance.

{\small
\bibliographystyle{ieee_fullname}
\bibliography{egbib}

\begin{thebibliography}{10}\itemsep=-1pt

\bibitem{abbatematteo2019learning}
Ben Abbatematteo, Stefanie Tellex, and George Konidaris.
\newblock Learning to generalize kinematic models to novel objects.
\newblock In {\em CoRL}, pages 1289--1299, 2019.

\bibitem{besl1992method}
Paul~J Besl and Neil~D McKay.
\newblock Method for registration of 3-d shapes.
\newblock In {\em Sensor fusion IV: control paradigms and data structures},
  volume 1611, pages 586--606. International Society for Optics and Photonics,
  1992.

\bibitem{bohg2014robot}
Jeannette Bohg, Javier Romero, Alexander Herzog, and Stefan Schaal.
\newblock Robot arm pose estimation through pixel-wise part classification.
\newblock In {\em 2014 IEEE International Conference on Robotics and Automation
  (ICRA)}, pages 3143--3150. IEEE, 2014.

\bibitem{byravan2017se3}
Arunkumar Byravan and Dieter Fox.
\newblock Se3-nets: Learning rigid body motion using deep neural networks.
\newblock In {\em 2017 IEEE International Conference on Robotics and Automation
  (ICRA)}, pages 173--180. IEEE, 2017.

\bibitem{chen2019self}
Yuhua Chen, Cordelia Schmid, and Cristian Sminchisescu.
\newblock Self-supervised learning with geometric constraints in monocular
  video: Connecting flow, depth, and camera.
\newblock In {\em Proceedings of the IEEE international conference on computer
  vision (ICCV)}, pages 7063--7072, 2019.

\bibitem{daniele2020multiview}
Andrea~F Daniele, Thomas~M Howard, and Matthew~R Walter.
\newblock A multiview approach to learning articulated motion models.
\newblock In {\em Robotics Research}, pages 371--386. Springer, 2020.

\bibitem{dempster1977maximum}
Arthur~P Dempster, Nan~M Laird, and Donald~B Rubin.
\newblock Maximum likelihood from incomplete data via the em algorithm.
\newblock {\em Journal of the Royal Statistical Society: Series B
  (Methodological)}, 39(1):1--22, 1977.

\bibitem{desingh2019factored}
Karthik Desingh, Shiyang Lu, Anthony Opipari, and Odest~Chadwicke Jenkins.
\newblock Factored pose estimation of articulated objects using efficient
  nonparametric belief propagation.
\newblock In {\em 2019 International Conference on Robotics and Automation
  (ICRA)}, pages 7221--7227. IEEE, 2019.

\bibitem{fischler1981random}
Martin~A Fischler and Robert~C Bolles.
\newblock Random sample consensus: a paradigm for model fitting with
  applications to image analysis and automated cartography.
\newblock {\em Communications of the ACM}, 24(6):381--395, 1981.

\bibitem{godard2019digging}
Cl{\'e}ment Godard, Oisin Mac~Aodha, Michael Firman, and Gabriel~J Brostow.
\newblock Digging into self-supervised monocular depth estimation.
\newblock In {\em Proceedings of the IEEE international conference on computer
  vision (ICCV)}, pages 3828--3838, 2019.

\bibitem{gulde2019ropose}
Thomas Gulde, Dennis Ludl, Johann Andrejtschik, Salma Thalji, and Crist{\'o}bal
  Curio.
\newblock Ropose-real: real world dataset acquisition for data-driven
  industrial robot arm pose estimation.
\newblock In {\em 2019 International Conference on Robotics and Automation
  (ICRA)}, pages 4389--4395. IEEE, 2019.

\bibitem{gulde2018ropose}
Thomas Gulde, Dennis Ludl, and Crist{\'o}bal Curio.
\newblock Ropose: Cnn-based 2d pose estimation of industrial robots.
\newblock In {\em 2018 IEEE 14th International Conference on Automation Science
  and Engineering (CASE)}, pages 463--470. IEEE, 2018.

\bibitem{ha2020learning}
Huy Ha, Jingxi Xu, and Shuran Song.
\newblock Learning a decentralized multi-arm motion planner.
\newblock {\em arXiv preprint arXiv:2011.02608}, 2020.

\bibitem{hartley2003multiple}
Richard Hartley and Andrew Zisserman.
\newblock {\em Multiple view geometry in computer vision}.
\newblock Cambridge university press, 2003.

\bibitem{ilg2017flownet}
Eddy Ilg, Nikolaus Mayer, Tonmoy Saikia, Margret Keuper, Alexey Dosovitskiy,
  and Thomas Brox.
\newblock Flownet 2.0: Evolution of optical flow estimation with deep networks.
\newblock In {\em Proceedings of the IEEE conference on computer vision and
  pattern recognition}, pages 2462--2470, 2017.

\bibitem{jain2020screwnet}
Ajinkya Jain, Rudolf Lioutikov, and Scott Niekum.
\newblock Screwnet: Category-independent articulation model estimation from
  depth images using screw theory.
\newblock {\em arXiv preprint arXiv:2008.10518}, 2020.

\bibitem{katz2008manipulating}
Dov Katz and Oliver Brock.
\newblock Manipulating articulated objects with interactive perception.
\newblock In {\em 2008 IEEE International Conference on Robotics and Automation
  (ICRA)}, pages 272--277. IEEE, 2008.

\bibitem{katz2013interactive}
Dov Katz, Moslem Kazemi, J~Andrew Bagnell, and Anthony Stentz.
\newblock Interactive segmentation, tracking, and kinematic modeling of unknown
  3d articulated objects.
\newblock In {\em 2013 IEEE International Conference on Robotics and Automation
  (ICRA)}, pages 5003--5010. IEEE, 2013.

\bibitem{kumar2016spatiotemporal}
Suren Kumar, Vikas Dhiman, Madan~Ravi Ganesh, and Jason~J Corso.
\newblock Spatiotemporal articulated models for dynamic slam.
\newblock {\em arXiv preprint arXiv:1604.03526}, 2016.

\bibitem{lambrecht2019towards}
Jens Lambrecht and Linh K{\"a}stner.
\newblock Towards the usage of synthetic data for marker-less pose estimation
  of articulated robots in rgb images.
\newblock In {\em 2019 19th International Conference on Advanced Robotics
  (ICAR)}, pages 240--247. IEEE, 2019.

\bibitem{lee2020camera}
Timothy~E Lee, Jonathan Tremblay, Thang To, Jia Cheng, Terry Mosier, Oliver
  Kroemer, Dieter Fox, and Stan Birchfield.
\newblock Camera-to-robot pose estimation from a single image.
\newblock In {\em 2020 IEEE International Conference on Robotics and Automation
  (ICRA)}, pages 9426--9432. IEEE, 2020.

\bibitem{li2020category}
Xiaolong Li, He Wang, Li Yi, Leonidas~J Guibas, A~Lynn Abbott, and Shuran Song.
\newblock Category-level articulated object pose estimation.
\newblock In {\em Proceedings of the IEEE/CVF Conference on Computer Vision and
  Pattern Recognition (CVPR)}, pages 3706--3715, 2020.

\bibitem{loper2015smpl}
Matthew Loper, Naureen Mahmood, Javier Romero, Gerard Pons-Moll, and Michael~J
  Black.
\newblock Smpl: A skinned multi-person linear model.
\newblock {\em ACM transactions on graphics (TOG)}, 34(6):1--16, 2015.

\bibitem{ma2020edge}
Qun Ma, Jianwei Niu, Zhenchao Ouyang, Mo Li, Tao Ren, and QingFeng Li.
\newblock Edge computing-based 3d pose estimation and calibration for robot
  arms.
\newblock In {\em 2020 7th IEEE International Conference on Cyber Security and
  Cloud Computing (CSCloud)/2020 6th IEEE International Conference on Edge
  Computing and Scalable Cloud (EdgeCom)}, pages 246--251. IEEE, 2020.

\bibitem{magrini2020human}
Emanuele Magrini, Federica Ferraguti, Andrea~Jacopo Ronga, Fabio Pini,
  Alessandro De~Luca, and Francesco Leali.
\newblock Human-robot coexistence and interaction in open industrial cells.
\newblock {\em Robotics and Computer-Integrated Manufacturing}, 61:101846,
  2020.

\bibitem{martin2016integrated}
Roberto Mart{\'\i}n-Mart{\'\i}n, Sebastian H{\"o}fer, and Oliver Brock.
\newblock An integrated approach to visual perception of articulated objects.
\newblock In {\em 2016 IEEE International Conference on Robotics and Automation
  (ICRA)}, pages 5091--5097. IEEE, 2016.

\bibitem{michel2015pose}
Frank Michel, Alexander Krull, Eric Brachmann, Michael~Ying Yang, Stefan
  Gumhold, and Carsten Rother.
\newblock Pose estimation of kinematic chain instances via object coordinate
  regression.
\newblock In {\em BMVC}, pages 181--1, 2015.

\bibitem{miseikis2018multi}
Justinas Miseikis, Inka Brijacak, Saeed Yahyanejad, Kyrre Glette, Ole~Jakob
  Elle, and Jim Torresen.
\newblock Multi-objective convolutional neural networks for robot localisation
  and 3d position estimation in 2d camera images.
\newblock In {\em 2018 15th International Conference on Ubiquitous Robots
  (UR)}, pages 597--603. IEEE, 2018.

\bibitem{ortenzi2018vision}
Valerio Ortenzi, Naresh Marturi, Michael Mistry, Jeffrey Kuo, and Rustam
  Stolkin.
\newblock Vision-based framework to estimate robot configuration and kinematic
  constraints.
\newblock {\em IEEE/ASME Transactions on Mechatronics}, 23(5):2402--2412, 2018.

\bibitem{ortenzi2016vision}
Valerio Ortenzi, Naresh Marturi, Rustam Stolkin, Jeffrey~A Kuo, and Michael
  Mistry.
\newblock Vision-guided state estimation and control of robotic manipulators
  which lack proprioceptive sensors.
\newblock In {\em 2016 IEEE/RSJ International Conference on Intelligent Robots
  and Systems (IROS)}, pages 3567--3574. IEEE, 2016.

\bibitem{pavlasek2020parts}
Jana Pavlasek, Stanley Lewis, Karthik Desingh, and Odest~Chadwicke Jenkins.
\newblock Parts-based articulated object localization in clutter using belief
  propagation.
\newblock {\em arXiv preprint arXiv:2008.02881}, 2020.

\bibitem{schmidt2015depth}
Tanner Schmidt, Katharina Hertkorn, Richard Newcombe, Zoltan Marton, Michael
  Suppa, and Dieter Fox.
\newblock Depth-based tracking with physical constraints for robot
  manipulation.
\newblock In {\em 2015 IEEE International Conference on Robotics and Automation
  (ICRA)}, pages 119--126. IEEE, 2015.

\bibitem{schmidt2014dart}
Tanner Schmidt, Richard~A Newcombe, and Dieter Fox.
\newblock Dart: Dense articulated real-time tracking.
\newblock In {\em Robotics: Science and Systems}, volume~2. Berkeley, CA, 2014.

\bibitem{szeliski2010computer}
Richard Szeliski.
\newblock {\em Computer vision: algorithms and applications}.
\newblock Springer Science \& Business Media, 2010.

\bibitem{wang2019normalized}
He Wang, Srinath Sridhar, Jingwei Huang, Julien Valentin, Shuran Song, and
  Leonidas~J Guibas.
\newblock Normalized object coordinate space for category-level 6d object pose
  and size estimation.
\newblock In {\em Proceedings of the IEEE Conference on Computer Vision and
  Pattern Recognition (CVPR)}, pages 2642--2651, 2019.

\bibitem{widmaier2016robot}
Felix Widmaier, Daniel Kappler, Stefan Schaal, and Jeannette Bohg.
\newblock Robot arm pose estimation by pixel-wise regression of joint angles.
\newblock In {\em 2016 IEEE International Conference on Robotics and Automation
  (ICRA)}, pages 616--623. IEEE, 2016.

\bibitem{xiang2020sapien}
Fanbo Xiang, Yuzhe Qin, Kaichun Mo, Yikuan Xia, Hao Zhu, Fangchen Liu, Minghua
  Liu, Hanxiao Jiang, Yifu Yuan, He Wang, et~al.
\newblock Sapien: A simulated part-based interactive environment.
\newblock In {\em Proceedings of the IEEE/CVF Conference on Computer Vision and
  Pattern Recognition (CVPR)}, pages 11097--11107, 2020.

\bibitem{ye2014real}
Mao Ye and Ruigang Yang.
\newblock Real-time simultaneous pose and shape estimation for articulated
  objects using a single depth camera.
\newblock In {\em Proceedings of the IEEE Conference on Computer Vision and
  Pattern Recognition (CVPR)}, pages 2345--2352, 2014.

\bibitem{yi2018deep}
Li Yi, Haibin Huang, Difan Liu, Evangelos Kalogerakis, Hao Su, and Leonidas
  Guibas.
\newblock Deep part induction from articulated object pairs.
\newblock {\em arXiv preprint arXiv:1809.07417}, 2018.

\bibitem{zhou20193d}
Fan Zhou, Zijing Chi, Chungang Zhuang, and Han Ding.
\newblock 3d pose estimation of robot arm with rgb images based on deep
  learning.
\newblock In {\em International Conference on Intelligent Robotics and
  Applications}, pages 541--553. Springer, 2019.

\bibitem{zuo2019craves}
Yiming Zuo, Weichao Qiu, Lingxi Xie, Fangwei Zhong, Yizhou Wang, and Alan~L
  Yuille.
\newblock Craves: Controlling robotic arm with a vision-based economic system.
\newblock In {\em Proceedings of the IEEE Conference on Computer Vision and
  Pattern Recognition (CVPR)}, pages 4214--4223, 2019.

\end{thebibliography}
}

\end{document}